%% file: main.tex
\pdfoutput=1

\documentclass[11pt]{article}

\usepackage{latex/acl}

\usepackage{times}
\usepackage{latexsym}

\usepackage[T1]{fontenc}

\usepackage[utf8]{inputenc}

\usepackage{microtype}

\usepackage{inconsolata}

\usepackage{booktabs}
\usepackage{graphicx}
\usepackage{graphics}
\usepackage{amsmath}
\usepackage{multirow}
\usepackage{paralist}
\usepackage{color,soul}

\usepackage[normalem]{ulem}

\title{Improving Image Captioning by Mimicking Human Reformulation Feedback at Inference-time}

\author{Uri Berger$^{1,2}$, Omri Abend$^1$, Lea Frermann$^2$ and Gabriel Stanovsky$^1$ \\
$^1$School of Computer Science and Engineering, The Hebrew University of Jerusalem \\
$^2$School of Computing and Information Systems, University of Melbourne \\
\texttt{\{uri.berger2, gabriel.stanovsky, omri.abend\}@mail.huji.ac.il} \\
\texttt{lea.frermann@unimelb.edu.au}}

\begin{document}
\maketitle
\begin{abstract}
\input{sections/00_abstract}
\end{abstract}

\section{Introduction}
\input{sections/01_introduction}

\section{Related Work}
\input{sections/02_background}

\section{Modeling Reformulation Feedback} \label{sec:reformulation_feedback}
\input{sections/03_reformulation_feedback}

\section{Reformulation for Improved Factuality} \label{sec:error_correction}
\input{sections/04_error_correction}

\section{Reformulation for Style Transfer} \label{sec:style_transfer}
\input{sections/05_stylized_captioning}

\section{Discussion}
\input{sections/06_discussion}

\section*{Limitations}
\input{sections/limitations}

\section*{Ethics Statement}
\input{sections/ethics_statement}

\section*{Acknowledgements}
We would like to thank the anonymous reviewers for their helpful comments and feedback.
We would also like to thank Desmond Elliott for consulting.
This work was supported in part by the HUJI-UoM joint PhD program.

\bibliography{anthology,custom}

\appendix

\section{Model Training Details}
\label{sec:app_training}
\input{sections/app_a_training_details}

\section{Data Collection} \label{sec:app_data_collection}
\input{sections/app_b_data_collection}

\section{Used Packages}
\input{sections/app_c_used_packages}

\section{More Examples} \label{sec:app_more_examples}
\input{sections/app_d_more_examples}

\section{Analysis of BLIP Reformulation} \label{sec:app_blip_analysis}
\input{sections/app_e_blip_analysis}

\end{document}

%% file: sections/00_abstract.tex
Incorporating automatically predicted human feedback into the process of training generative models has attracted substantial recent interest, while {\it feedback at inference time} has received less attention.
The typical feedback at training time, i.e.,
preferences of choice given two samples, does not naturally transfer to the inference phase.
We introduce a novel type of feedback -- caption reformulations -- and train models to mimic reformulation feedback based on human annotations.
Our method does not require training the image captioning model itself, thereby demanding substantially less computational effort.
We experiment with two types of reformulation feedback: 
first, we collect a dataset of human reformulations that correct errors in the generated captions. We find that incorporating reformulation models trained on this data into the inference phase of existing image captioning models results in
improved captions, especially when the original captions are of low quality. We apply our method to non-English image captioning, a domain where robust models are less prevalent, and gain substantial improvement.
Second, we apply reformulations to style transfer. Quantitative evaluations reveal state-of-the-art performance on German image captioning and English style transfer, while human validation with a detailed comparative framework exposes the specific axes of
improvement.\footnote{Our code and data are available here: \\ \href{https://github.com/uriberger/re_cap.git}{github.com/uriberger/re\_cap.git}}

%% file: sections/01_introduction.tex
\begin{figure} [tb]
    \centering
    \includegraphics[width=6cm]{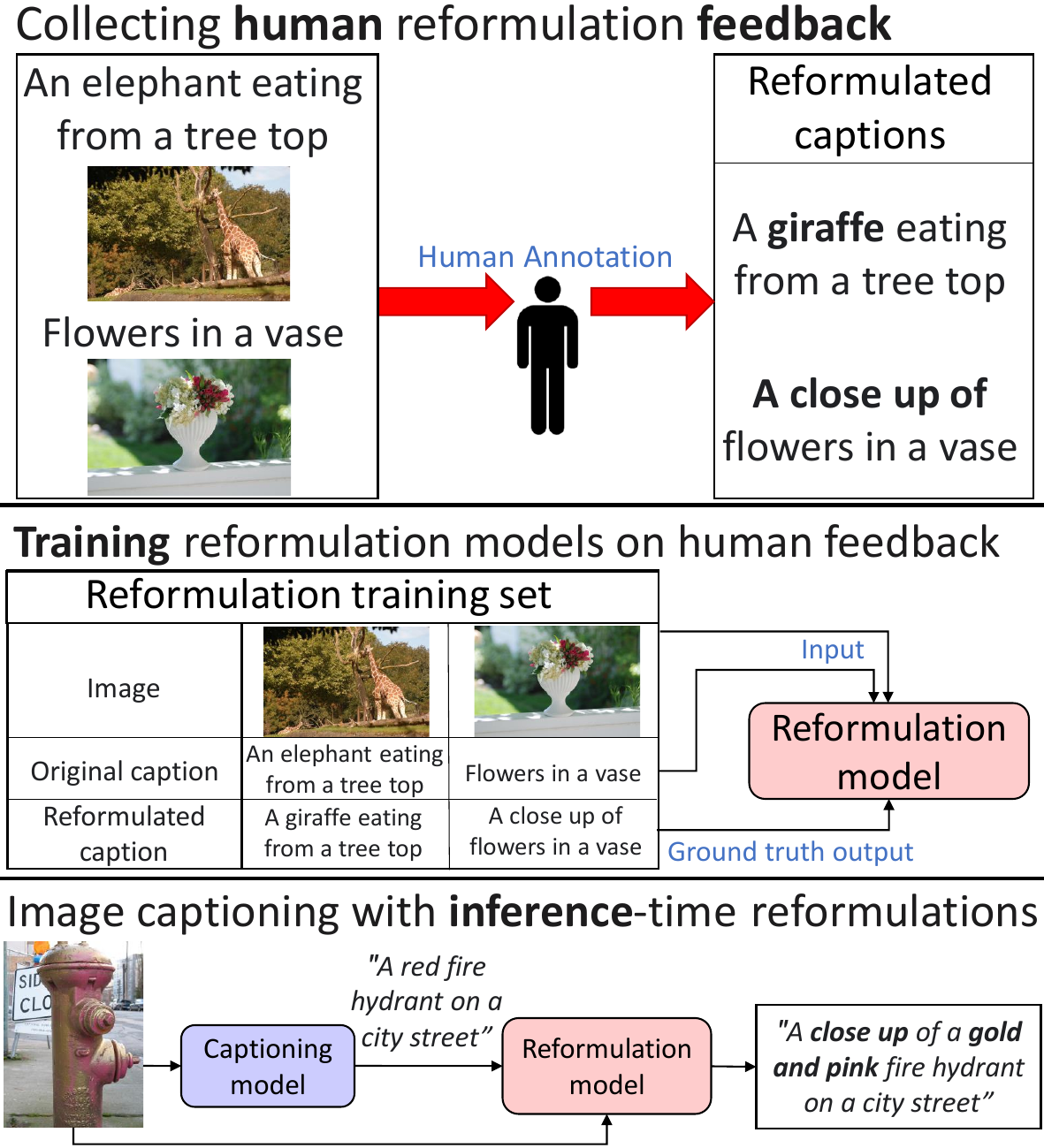}
    \caption{Our proposed method with reformulation for improved factuality as an example. Top: Collecting human-written reformulations of model captions. Center: Using the collected data to train models to generate reformulations, given an input image and original caption. Bottom: Combining an off-the-shelf captioning model (no training) with our reformulation model, to adapt generated captions at inference time.}
    \label{fig:pipeline}
\end{figure}

There is a growing interest in feedback models that approximate human feedback \emph{during training} of generative models.
While resulting generative models achieve improved performance on automatic metrics and human evaluations~\citep{ouyang2022training, faltings2023interactive}, the use of feedback models during training requires the generative model to be trained or at least fine-tuned.

The use of such feedback models \emph{during inference} poses no such requirement, but was nevertheless generally overlooked by previous studies. One reason for this limited interest is the type of feedback that existing feedback models predict:
comparative feedback, e.g., by predicting human preference for one of two generated candidate outputs~\citep[as used in Reinforcement Learning from Human Feedback,][]{stiennon2020learning}. While this type of feedback naturally translates into a reward function to be used during training, it is less clear how to employ it at inference time when model parameters are fixed.

We bridge this gap by proposing a novel type of feedback, namely \textbf{reformulation} (see Figure~\ref{fig:pipeline} and Section~\ref{sec:reformulation_feedback}).
We focus on the image captioning task, since it provides a good testing ground for adapting a general model to fit specific user intent. For example, one user may require the captions to describe the colors in the image while another would focus on a specific style of generated captions.

When providing reformulation feedback for model-generated captions, human annotators receive an image and a model-generated textual description as input, and subsequently produce text that is as similar as possible to the input text but also incorporates an additional desired attribute, e.g., improved factuality or a desired style (Fig.~\ref{fig:pipeline}, top). We train models to mimic this type of feedback (Fig.~\ref{fig:pipeline}, center) and integrate them into the inference phase of off-the-shelf image captioning models (Fig.~\ref{fig:pipeline}, bottom).

A small amount of data (a few thousand samples, as demonstrated in our experiments in Sections~\ref{sec:error_correction} and \ref{sec:style_transfer}) is sufficient to train a reformulation model that once trained, can be applied to any captioning model without further training it, making reformulation models a much more efficient alternative to training-time feedback models that require to retrain the captioning model.

To study the benefits of this type of feedback, we focus on two reformulation attributes. First, we train models to rewrite the input caption with improved factuality (Section~\ref{sec:error_correction}).
We collect English reformulation data by asking human annotators to correct errors in generated captions while making minimal changes, and use this data to train a reformulation model.
We then use the reformulation model on captions generated by off-the-shelf English models. We show that the automatic reformulation process notably improves captions generated by weaker models, while careful analysis including a fine-grained human evaluation paradigm reveals that, similar to human reformulations, the most notable factor in the improvement of the automatic reformulation process is adding missing information.
To further investigate the utility of our method in domains where existing models are weak (``challenge domains''\footnote{We define ``challenge domains'' as the (many) domains dominated by weaker models, e.g., low-resource scenarios or niche domains less amenable to established model architectures.}), we propose a cross-lingual pipeline for reformulation in German image captioning and show notable improvement, achieving state-of-the-art performance in German image captioning on the Multi30k~\cite{elliott2016multi30k} dataset.

Second, we cast caption style transfer as a reformulation task (Section~\ref{sec:style_transfer}). We use existing parallel stylized and non-stylized caption data, and train a reformulation model to preserve the structure of the input caption while adapting its style to a given target.  
We build on powerful but style-agnostic captioning models using style-reformulation at inference time, achieving state-of-the-art performance on the FlickrStyle dataset on automatic metrics, while our human evaluation paradigm confirms that the reformulated captions are more stylized than competitive baselines.

Each of the reformulation attributes studied in this work (improved factuality and style transfer) reveal a use case for our method. In the first, the goal of the user is to improve captioning models in challenge domains. This is accomplished by selecting a reformulation attribute that will improve the quality of the captions (factuality in our case) and apply corresponding reformulation models to a weak captioning model. In the second case the user aspires to generate high-quality captions in a specific style, and therefore utilizes a robust captioning model to generate high quality captions and then change their style using reformulation.

%% file: sections/02_background.tex
We classify related work by the type of feedback (human/model-generated) and the phase in which the feedback is applied (training/inference): human feedback during training (Section~\ref{sec:bg_raw_feedback}), model-generated feedback during training (Section~\ref{sec:bg_model_feedback_training}), and model-generated feedback during inference (Section~\ref{sec:bg_model_feedback_inference}, our study is included in this category).
  
\subsection{Human Feedback during Training} 
\label{sec:bg_raw_feedback}

A large volume of previous studies collect human feedback and use it directly to improve training. Most studies focus on either comparisons (which of two candidate texts is better) or ratings as a reward signal in reinforcement learning for various tasks, e.g., dialogue~\citep{jaques-etal-2020-human}, machine translation~\citep{kreutzer-etal-2018-neural} or semantic parsing~\citep{lawrence-riezler-2018-improving}. \citet{kreutzer-etal-2020-correct} fine-tune a generative model with on-line feedback from human annotators.

Other types of training-time feedback include natural language comments~\citep{campos2022training}. Most similar to our reformulation feedback are edits~\citep{liu2022second, lu-etal-2023-towards}, where humans change an incorrect response generated by dialog models into a correct response.
While these studies use feedback directly we train models to predict it, avoiding the necessity to collect annotations each time the feedback is used.

\paragraph{Image captioning.}
Several previous studies trained captioning models with raw human feedback. \citet{Shen_2019_ICCV} propose a model that generates a caption and subsequently generates a question pertaining to factual information within that caption. This question is then answered by a human. No questions are generated during inference.
\citet{seo2020reinforcing} use human ratings of captions as rewards in a reinforcement learning framework.

\citet{ling2017teaching} are most similar to our study: they compare training captioning models on human-generated captions with training on reformulations, showing that the latter improves standard metrics.
However, they use human-generated reformulations during training while we use model-generated reformulations during inference.

\subsection{Model-Generated Feedback during Training} 
\label{sec:bg_model_feedback_training}
Several works train feedback models, but use these models during training, again predominantly focusing on comparisons or ratings feedback. Early studies~\citep{christiano2017deep, ibarz2018reward} use feedback models to train agents in simulated environments and games. Others use feedback models to train language models for specific tasks such as summarization~\citep{ziegler2019fine, stiennon2020learning}, machine translation~\citep{kreutzer-etal-2018-reliability} and visual storytelling~\citep{hsu-etal-2021-plot}. Recently, feedback models were used in training of general-purpose large language models~\citep[e.g. GPT-4,][]{OpenAI2023GPT4TR}. Most relatedly, \citet{faltings2023interactive} investigate reformulation feedback models, but only during training.
Finally, constitutional AI~\cite{bai2022constitutional} use similar ideas to train non-harmful models but use model feedback rather than human feedback.

\subsection{Model-Generated Feedback during Inference}
\label{sec:bg_model_feedback_inference}

Most similar to ours, some previous studies apply feedback models at inference time. \citet{hsu-etal-2019-visual} train models to predict human post-edits of model generated text but focus on the visual storytelling task.
\citet{ramos2023aligning} apply metrics trained to predict human rating feedback for reranking model outputs in machine translation.
To the best of our knowledge, we are the first to use feedback models at inference time for image captioning.

%% file: sections/03_reformulation_feedback.tex

In this Section, we define our notion of reformulation feedback.
A human annotator observes an image and a caption describing the image, and produces a caption that 1) incorporates some desired attribute (e.g., factuality or some desired style) , and 2) is as similar to the input caption as possible.
Since these two criteria are in conflict we emphasize that the first requirement is obligatory, but annotators should make minimal changes to achieve it.
Note that reformulation may be applied in any generation task, but here we focus only on image captioning.

In this study we focus on two attributes of reformulation feedback: improved factuality (Section~\ref{sec:error_correction}) and style transfer (Section~\ref{sec:style_transfer}). 

\paragraph{Reformulation model.}
Recent research examined frameworks of multimodal input (image+text) and unimodal output (text), demonstrating that fine-tuning a checkpoint that was pre-trained on general Vision-and-Language tasks is an effective approach~\citep[e.g., in Visual Qustion Answering,][]{chen2022pali}. We follow this strategy by fine-tuning the pre-trained mPLUG~\citep{li2022mplug} checkpoint\footnote{We also experimented with BLIP, but mPLUG performed significantly better.} on reformulation data\footnote{For the full training details see Appendix~\ref{sec:app_training}.}.

%% file: sections/04_error_correction.tex
Captioning models in challenge domains, e.g., non-English captioning, tend to generate captions of lower quality compared to English captioning models. We propose to use reformulations to improve the factuality of models in these domains. In this Section, we study this use case.
We first describe data collection and then apply our model to English and German image captioning.

\subsection{Data Collection} \label{sec:data_collection}

\paragraph{Data.}
To generate an initial set of image captions, we use three publicly available captioning models, that vary in architecture, size and amounts of training data: BLIP~\citep{li2022blip}, mPLUG~\citep{li2022mplug}, and ClipCap~\citep{mokady2021clipcap}. We randomly sample 1405 images from the test sets of MSCOCO~\cite{lin2014microsoft} and Flickr30k~\cite{young2014image}, and generate a caption with each model.

\paragraph{Annotation.}
Human annotators were shown an image and a model-generated caption, and asked to reformulate the caption so that (a) it is as similar as possible to the original caption and (b) any errors in the original caption are corrected (if any errors were present).

Annotators were instructed to consider a wide range of errors in their feedback, including hallucinations (describing elements that are not present in the image), partial descriptions (failing to describe a key element in the image) and replacements (using an incorrect word to describe an element in the image).

We use Amazon Mechanical Turk to recruit annotators. For the full details on annotator recruitment, guidelines and payment, see Appendix~\ref{sec:app_data_collection}.

\paragraph{Data analysis.}
In 864 samples (16.6\%) the annotators chose not to change the original caption. The mean Levenshtein distance\footnote{Minimum number of words needed to be added, removed or replaced to get from original to reformulated caption.} is 4.79. Additionally, we sample 100 random captions that were changed by the annotators and classify the changes to the element changed (object, action, object attribute, setting, other) and the nature of the change (add, replace, remove, rewrite). {`Setting}' changes are changes in the setting of the caption (e.g., adding the location in which the caption takes place is classified as {`add setting'}). `Other' captures any change that is not covered by the first four elements. 
If most of the objects, actions and attributes in the reformulated caption differ from those of the original caption, we classify the change as {`rewrite'}. Results in Table~\ref{tab:re_dataset_analysis} show that \emph{object} is the most frequently changed element: in 51\% of the captions an object was added, replaced or removed. The most common type of change (applied in 73\% of the captions) is adding information.
We find that all the annotators' modifications were valid\footnote{For the list of manually examined captions, see supplementary materials.}.

\input{tables_and_plots/reformulation_dataset_analysis}

\subsection{Improved Factuality for English Image Captioning} \label{sec:english}

\input{sections/04.2_english}

\subsection{Improved Factuality for Cross-Lingual Image Captioning} \label{sec:german}

\input{sections/04.3_german}

%% file: tables_and_plots/reformulation_dataset_analysis.tex
\begin{table}[!t]
\small
\centering
\resizebox{7.3cm}{!}{%
\begin{tabular}{lccccc}
\toprule
 & Add & Replace & Remove & Rewrite & Total \\
\midrule
Object & 24 & 24 & 3 & -- & 51 \\
Action & 11 & 7 & 0 & -- & 18 \\
Attribute & 12 & 0 & 3 & -- & 15 \\
Setting & 26 & 3 & 0 & -- & 29 \\
Other & 0 & 9 & 0 & -- & 9 \\
Total & 73 & 43 & 6 & 15 & \\
\bottomrule
\end{tabular}%
}
\caption{Statistics for reformulations of 100 random labeled data points. One reformulation may contain several operations.}
\label{tab:re_dataset_analysis}
\end{table}

%% file: sections/04.2_english.tex
In this section we experiment on English data.
We use off-the-shelf captioning models on well known captioning datasets and reformulate the generated captions using the model described in Section~\ref{sec:reformulation_feedback} trained on the data described in Section~\ref{sec:data_collection}.
To test the reformulation model on data both from a familiar and an unfamiliar distribution, we use the models (BLIP, ClipCap, mPLUG) and datasets (MSCOCO, Flickr30k) used to generate the reformulation training data (Section~\ref{sec:data_collection}) excluding the images that were already presented to the reformulation model during training, as well as models (GIT: \citealt{wang2022git}, vit\_gpt2: \citealt{kumar2022imagecaptioning}) and datasets~\citep[XM3600:][]{thapliyal2022crossmodal} with which the reformulation model is unfamiliar.

As described above, in this use case we expect to improve the factuality of weaker models. We therefore mainly focus on relatively weak captioning models: we use the pretrained only (not finetuned) checkpoint of mPLUG, the base version of GIT, and ClipCap and vit\_gpt2 which are realtively old and small models. For completion we also use one strong model, the finetuned checkpoint of BLIP.

\subsubsection{Automatic Evaluation} \label{sec:english_auto_eval}

We present the change in performance for different metrics in Table~\ref{tab:re_during_inference}. We use the commonly used~\citep[e.g.,][]{li2022mplug, li2022blip} metrics BLEU-4~\cite{papineni2002bleu}, METEOR~\cite{banerjee-lavie-2005-meteor} CIDEr~\cite{vedantam2015cider}, and SPICE~\cite{anderson2016spice}. In addition to these 4 general metrics we report the performance on different types of sentential elements, provided by the SPICE metric
(objects, relations, attributes, size words, color words, cardinality words).
This allows us to observe a change in performance specifically regarding the sentential elements that our reformulation models are trained to address (Table~\ref{tab:re_dataset_analysis}).

\input{tables_and_plots/experiments/re_during_inference}

For the weaker models (mPLUG, ClipCap, GIT, vit\_gtp2) we see an improvement across all datasets and general metrics. For BLIP we see a minor decrease in performance on BLEU-4 and CIDEr, and a minor increase on METEOR and SPICE.
Turning to SPICE components, improvement was observed across all weaker models, datasets and components, except for Color with GIT on Flickr30k. For BLIP, improvement was observed in most configurations, but most notably in color words.
Therefore, our reformulation model is particularly well-suited for domains characterized by a lack of robust models, such as non-English image captioning, on which we focus in Section~\ref{sec:german}.

\begin{figure*} [tb]
    \centering
    \includegraphics[width=14cm]{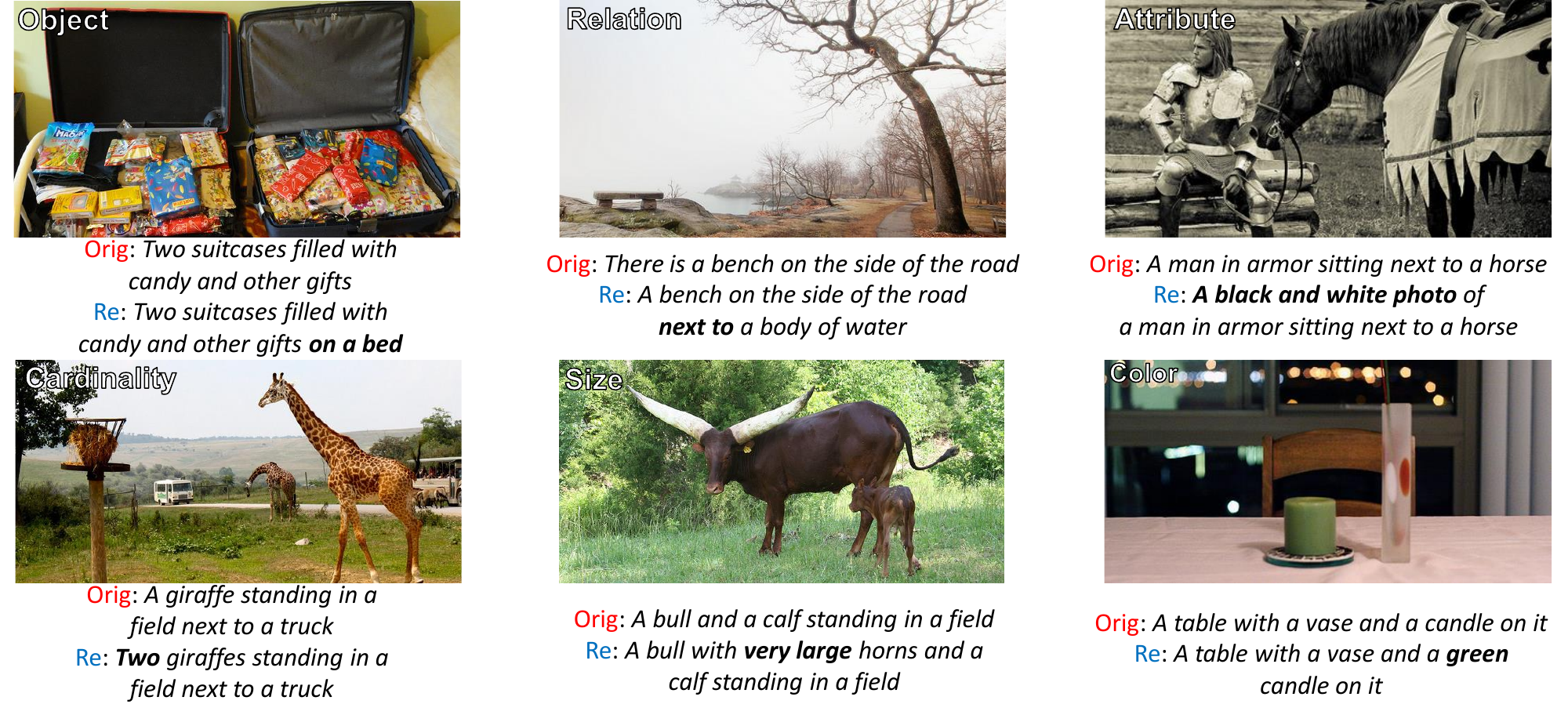}
    \caption{Examples in which the reformulated captions achieved better results than the original ones, in all SPICE elements. Orig: the caption generated by the model. Re: the reformulated caption.}
    \label{fig:spice_improvement}
\end{figure*}

Figure~\ref{fig:spice_improvement} shows examples where SPICE scores were notably higher after reformulation, for each SPICE element. In accordance with Table~\ref{tab:re_dataset_analysis}, most of the improvement originates from information that was added during reformulation.

\subsubsection{Human Evaluation} \label{sec:english_human_eval}
To qualitatively evaluate the changes during reformulation, we propose a fine-grained human evaluation paradigm. For each of the models we randomly sample images from each of the datasets (17 from MSCOCO and Flickr30k, 16 from XM3600 to a total of 50 per model), and present human annotators with the images along with the original caption and the reformulated caption, in randomly shuffled order and without indicating the source of each caption.
As we expect to observe notable improvement for weaker models, we exclude BLIP from this analysis\footnote{See Appendix~\ref{sec:app_blip_analysis} for a similar analysis for BLIP.}.
Three on-site annotators with high English proficiency assessed the captions. For each sample, the annotators  answer the following questions, each related to one axis of caption quality (in bold):
\begin{compactitem}
    \setlength\itemsep{0em}
     \item \textbf{Faithfulness:} Which caption includes less content that is not in the image?
     \item \textbf{Completeness:} Which caption covers more elements of the image being described?
     \item \textbf{Accuracy:} Which caption uses fewer incorrect words to describe one of the object/activities in the image?
     \item \textbf{Detail:} Which caption includes more {\it properties} (such as color or shape) of the main objects in the image?
     \item \textbf{Overall:} Which caption is the better description of the image?
\end{compactitem}
For each question, the annotators were given three options (first caption is better, second is better, both are equal). If at least two annotators prefer one of the captions along an axis, we mark the caption as `better'. Otherwise both are considered `equal'.

\begin{figure} [tb]
    \centering
    \includegraphics[width=\columnwidth, height=3.9cm]{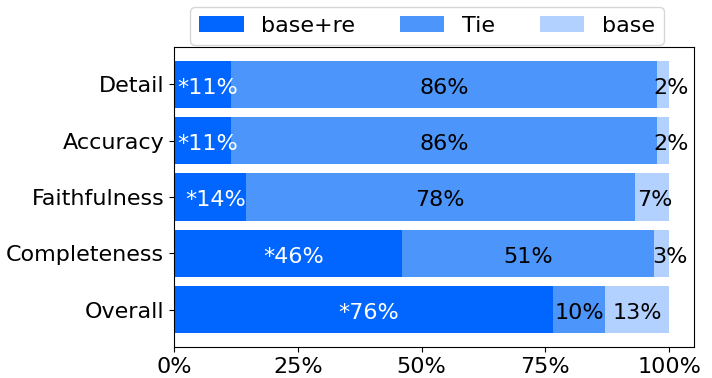}
    \caption{Results for human evaluation on different axes. We show proportions of preferences for generated captions without (base) and with (base+re) reformulations, and ties.
    * indicates a significant difference between base and base+re (Sign test; $p<0.05$).
    }
    \label{fig:english_human_eval}
\end{figure}

Figure~\ref{fig:english_human_eval} presents the results.
Across all axes, reformulated captions are significantly (Sign test, $p<0.05$) better than the original captions. Specifically, we see notable improvement in the overall quality (reformulated captions were better in 76\%) and the completeness (46\%) of the caption. This result is in line with the analysis presented in Table~\ref{tab:re_dataset_analysis}, where the most common feedback type was `addition' of information to the original caption.
The inter-annotator agreement using Fleiss' Kappa was $0.68, 0.68$ for completeness, overall \citep[substantial agreement,][]{landis1977measurement}, and $0.56, 0.55, 0.53$ for faithfulness, detail, accuracy (moderate agreement).

%% file: tables_and_plots/experiments/re_during_inference.tex
\definecolor{cellcolor1}{rgb}{0.9576, 1, 0.9576}
\definecolor{cellcolor2}{rgb}{0.9778, 1, 0.9778}
\definecolor{cellcolor3}{rgb}{0.9515, 1, 0.9515}
\definecolor{cellcolor4}{rgb}{0.9563, 1, 0.9563}
\definecolor{cellcolor5}{rgb}{0.8351, 1, 0.8351}
\definecolor{cellcolor6}{rgb}{0.8432, 1, 0.8432}
\definecolor{cellcolor7}{rgb}{0.784, 1, 0.784}
\definecolor{cellcolor8}{rgb}{0.7988, 1, 0.7988}
\definecolor{cellcolor9}{rgb}{0.4354, 1, 0.4354}
\definecolor{cellcolor10}{rgb}{0.71, 1, 0.71}
\definecolor{cellcolor11}{rgb}{0.5956, 1, 0.5956}
\definecolor{cellcolor12}{rgb}{0.6958, 1, 0.6958}
\definecolor{cellcolor13}{rgb}{0.9906, 1, 0.9906}
\definecolor{cellcolor14}{rgb}{0.9576, 1, 0.9576}
\definecolor{cellcolor15}{rgb}{0.9771, 1, 0.9771}
\definecolor{cellcolor16}{rgb}{0.9273, 1, 0.9273}
\definecolor{cellcolor17}{rgb}{1, 0.9133, 0.9133}
\definecolor{cellcolor18}{rgb}{0.996, 1, 0.996}
\definecolor{cellcolor19}{rgb}{1, 0.9467, 0.9467}
\definecolor{cellcolor20}{rgb}{0.9879, 1, 0.9879}
\definecolor{cellcolor21}{rgb}{0.8567, 1, 0.8567}
\definecolor{cellcolor22}{rgb}{0.9307, 1, 0.9307}
\definecolor{cellcolor23}{rgb}{0.7921, 1, 0.7921}
\definecolor{cellcolor24}{rgb}{0.8883, 1, 0.8883}
\definecolor{cellcolor25}{rgb}{0.7921, 1, 0.7921}
\definecolor{cellcolor26}{rgb}{0.8075, 1, 0.8075}
\definecolor{cellcolor27}{rgb}{0.6265, 1, 0.6265}
\definecolor{cellcolor28}{rgb}{0.7719, 1, 0.7719}
\definecolor{cellcolor29}{rgb}{0.6958, 1, 0.6958}
\definecolor{cellcolor30}{rgb}{0.7813, 1, 0.7813}
\definecolor{cellcolor31}{rgb}{0.8708, 1, 0.8708}
\definecolor{cellcolor32}{rgb}{0.8143, 1, 0.8143}
\definecolor{cellcolor33}{rgb}{0.86, 1, 0.86}
\definecolor{cellcolor34}{rgb}{0.8849, 1, 0.8849}
\definecolor{cellcolor35}{rgb}{0.7651, 1, 0.7651}
\definecolor{cellcolor36}{rgb}{0.8257, 1, 0.8257}
\definecolor{cellcolor37}{rgb}{1, 0.8917, 0.8917}
\definecolor{cellcolor38}{rgb}{0.9913, 1, 0.9913}
\definecolor{cellcolor39}{rgb}{1, 0.9633, 0.9633}
\definecolor{cellcolor40}{rgb}{0.9733, 1, 0.9733}
\definecolor{cellcolor41}{rgb}{0.9017, 1, 0.9017}
\definecolor{cellcolor42}{rgb}{0.9435, 1, 0.9435}
\definecolor{cellcolor43}{rgb}{0.8567, 1, 0.8567}
\definecolor{cellcolor44}{rgb}{0.9058, 1, 0.9058}
\definecolor{cellcolor45}{rgb}{0, 1, 0}
\definecolor{cellcolor46}{rgb}{0.6649, 1, 0.6649}
\definecolor{cellcolor47}{rgb}{0.5942, 1, 0.5942}
\definecolor{cellcolor48}{rgb}{0.7261, 1, 0.7261}
\definecolor{cellcolor49}{rgb}{0.6871, 1, 0.6871}
\definecolor{cellcolor50}{rgb}{0.8392, 1, 0.8392}
\definecolor{cellcolor51}{rgb}{0.8412, 1, 0.8412}
\definecolor{cellcolor52}{rgb}{0.9199, 1, 0.9199}
\definecolor{cellcolor53}{rgb}{0.7826, 1, 0.7826}
\definecolor{cellcolor54}{rgb}{0.8769, 1, 0.8769}
\definecolor{cellcolor55}{rgb}{0.7705, 1, 0.7705}
\definecolor{cellcolor56}{rgb}{0.8533, 1, 0.8533}
\definecolor{cellcolor57}{rgb}{1, 0.9417, 0.9417}
\definecolor{cellcolor58}{rgb}{0.9872, 1, 0.9872}
\definecolor{cellcolor59}{rgb}{0.9879, 1, 0.9879}
\definecolor{cellcolor60}{rgb}{0.9953, 1, 0.9953}
\definecolor{cellcolor61}{rgb}{0.9489, 1, 0.9489}
\definecolor{cellcolor62}{rgb}{0.9489, 1, 0.9489}
\definecolor{cellcolor63}{rgb}{0.896, 1, 0.896}
\definecolor{cellcolor64}{rgb}{0, 0.8428, 0}
\definecolor{cellcolor65}{rgb}{0.9879, 1, 0.9879}
\definecolor{cellcolor66}{rgb}{0.8668, 1, 0.8668}
\definecolor{cellcolor67}{rgb}{0.8237, 1, 0.8237}
\definecolor{cellcolor68}{rgb}{0.7544, 1, 0.7544}
\definecolor{cellcolor69}{rgb}{0.6534, 1, 0.6534}
\definecolor{cellcolor70}{rgb}{0, 0.7522, 0}
\definecolor{cellcolor71}{rgb}{0.3903, 1, 0.3903}
\definecolor{cellcolor72}{rgb}{0.3015, 1, 0.3015}
\definecolor{cellcolor73}{rgb}{0.7712, 1, 0.7712}
\definecolor{cellcolor74}{rgb}{0.3715, 1, 0.3715}
\definecolor{cellcolor75}{rgb}{0.5229, 1, 0.5229}
\definecolor{cellcolor76}{rgb}{0.609, 1, 0.609}
\definecolor{cellcolor77}{rgb}{0.6413, 1, 0.6413}
\definecolor{cellcolor78}{rgb}{0.7873, 1, 0.7873}
\definecolor{cellcolor79}{rgb}{0.959, 1, 0.959}
\definecolor{cellcolor80}{rgb}{0.8984, 1, 0.8984}
\definecolor{cellcolor81}{rgb}{0.7873, 1, 0.7873}
\definecolor{cellcolor82}{rgb}{0.6009, 1, 0.6009}
\definecolor{cellcolor83}{rgb}{0.8775, 1, 0.8775}
\definecolor{cellcolor84}{rgb}{0.1258, 1, 0.1258}
\definecolor{cellcolor85}{rgb}{0.9872, 1, 0.9872}
\definecolor{cellcolor86}{rgb}{0.9926, 1, 0.9926}
\definecolor{cellcolor87}{rgb}{0.9515, 1, 0.9515}
\definecolor{cellcolor88}{rgb}{0.9616, 1, 0.9616}
\definecolor{cellcolor89}{rgb}{0.9549, 1, 0.9549}
\definecolor{cellcolor90}{rgb}{0.8869, 1, 0.8869}
\definecolor{cellcolor91}{rgb}{0.9078, 1, 0.9078}
\definecolor{cellcolor92}{rgb}{0.8627, 1, 0.8627}
\definecolor{cellcolor93}{rgb}{0.747, 1, 0.747}
\definecolor{cellcolor94}{rgb}{0.2214, 1, 0.2214}
\definecolor{cellcolor95}{rgb}{0.9307, 1, 0.9307}
\definecolor{cellcolor96}{rgb}{0.7927, 1, 0.7927}
\definecolor{cellcolor97}{rgb}{0.7894, 1, 0.7894}
\definecolor{cellcolor98}{rgb}{0.8351, 1, 0.8351}
\definecolor{cellcolor99}{rgb}{0.4684, 1, 0.4684}
\definecolor{cellcolor100}{rgb}{0, 0.5, 0}
\definecolor{cellcolor101}{rgb}{0, 0.7792, 0}
\definecolor{cellcolor102}{rgb}{0.6635, 1, 0.6635}
\definecolor{cellcolor103}{rgb}{0.8385, 1, 0.8385}
\definecolor{cellcolor104}{rgb}{0, 0.9768, 0}
\definecolor{cellcolor105}{rgb}{0.8721, 1, 0.8721}
\definecolor{cellcolor106}{rgb}{0.6925, 1, 0.6925}
\definecolor{cellcolor107}{rgb}{0.4085, 0.7792, 0.4805}
\definecolor{cellcolor108}{rgb}{1, 0.795, 0.795}
\definecolor{cellcolor109}{rgb}{0.8721, 1, 0.8721}
\definecolor{cellcolor110}{rgb}{0.212, 1, 0.212}
\definecolor{cellcolor111}{rgb}{0.4879, 1, 0.4879}
\definecolor{cellcolor112}{rgb}{0.183, 1, 0.183}
\definecolor{cellcolor113}{rgb}{0.7645, 1, 0.7645}
\definecolor{cellcolor114}{rgb}{0, 0.96, 0}
\definecolor{cellcolor115}{rgb}{0.9899, 1, 0.9899}
\definecolor{cellcolor116}{rgb}{0.9852, 1, 0.9852}
\definecolor{cellcolor117}{rgb}{0.963, 1, 0.963}
\definecolor{cellcolor118}{rgb}{0.9441, 1, 0.9441}
\definecolor{cellcolor119}{rgb}{1, 0.94, 0.94}
\definecolor{cellcolor120}{rgb}{0.9024, 1, 0.9024}
\definecolor{cellcolor121}{rgb}{0.9152, 1, 0.9152}
\definecolor{cellcolor122}{rgb}{0.8984, 1, 0.8984}
\definecolor{cellcolor123}{rgb}{0.8183, 1, 0.8183}
\definecolor{cellcolor124}{rgb}{0.179, 1, 0.179}
\definecolor{cellcolor125}{rgb}{1, 1, 1}
\definecolor{cellcolor126}{rgb}{0.8802, 1, 0.8802}
\definecolor{cellcolor127}{rgb}{0.7526, 1, 0.7526}
\definecolor{cellcolor128}{rgb}{0.6427, 1, 0.6427}
\definecolor{cellcolor129}{rgb}{0.4643, 1, 0.4643}
\definecolor{cellcolor130}{rgb}{0, 0.5, 0}
\definecolor{cellcolor131}{rgb}{0.0316, 1, 0.0316}
\definecolor{cellcolor132}{rgb}{0.5828, 1, 0.5828}
\definecolor{cellcolor133}{rgb}{0.9233, 1, 0.9233}
\definecolor{cellcolor134}{rgb}{0.9684, 1, 0.9684}
\definecolor{cellcolor135}{rgb}{0.9105, 1, 0.9105}
\definecolor{cellcolor136}{rgb}{0.5236, 1, 0.5236}
\definecolor{cellcolor137}{rgb}{0.7732, 1, 0.7732}
\definecolor{cellcolor138}{rgb}{0.9913, 1, 0.9913}
\definecolor{cellcolor139}{rgb}{0.8816, 1, 0.8816}
\definecolor{cellcolor140}{rgb}{0.4798, 1, 0.4798}
\definecolor{cellcolor141}{rgb}{0.5437, 1, 0.5437}
\definecolor{cellcolor142}{rgb}{0.3991, 1, 0.3991}
\definecolor{cellcolor143}{rgb}{0.5108, 1, 0.5108}
\definecolor{cellcolor144}{rgb}{0.4004, 1, 0.4004}
\definecolor{cellcolor145}{rgb}{0.9939, 1, 0.9939}
\definecolor{cellcolor146}{rgb}{1, 0.9733, 0.9733}
\definecolor{cellcolor147}{rgb}{0.9751, 1, 0.9751}
\definecolor{cellcolor148}{rgb}{1, 0.8733, 0.8733}
\definecolor{cellcolor149}{rgb}{0.969, 1, 0.969}
\definecolor{cellcolor150}{rgb}{0.9596, 1, 0.9596}

\begin{table*}[!t]
\small
\centering
\resizebox{11cm}{!}{%
\begin{tabular}{cc|cccc|cccccc}
\toprule
\multirow{2}{*}{Dataset} & \multirow{2}{*}{Model} & \multicolumn{4}{c|}{General metrics} & \multicolumn{6}{c}{SPICE components} \\
& & B@4 & M & C & S & Obj & Rel & Att & Car & Siz & Col \\
\toprule
&ClipCap & \sethlcolor{cellcolor1}\hl{6.3} & \sethlcolor{cellcolor2}\hl{3.3} & \sethlcolor{cellcolor3}\hl{7.2} & \sethlcolor{cellcolor4}\hl{6.5} & \sethlcolor{cellcolor61}\hl{7.6} & \sethlcolor{cellcolor62}\hl{7.6} & \sethlcolor{cellcolor63}\hl{15.5} & \sethlcolor{cellcolor64}\hl{194.6} & \sethlcolor{cellcolor65}\hl{1.8} & \sethlcolor{cellcolor66}\hl{19.8} \\
&mPLUG & \sethlcolor{cellcolor5}\hl{24.5} & \sethlcolor{cellcolor6}\hl{23.3} & \sethlcolor{cellcolor7}\hl{32.1} & \sethlcolor{cellcolor8}\hl{29.9} & \sethlcolor{cellcolor67}\hl{26.2} & \sethlcolor{cellcolor68}\hl{36.5} & \sethlcolor{cellcolor69}\hl{51.5} & \sethlcolor{cellcolor70}\hl{221.1} & \sethlcolor{cellcolor71}\hl{90.6} & \sethlcolor{cellcolor72}\hl{44.8} \\
MSCOCO & GIT & \sethlcolor{cellcolor9}\hl{81.3} & \sethlcolor{cellcolor10}\hl{41.3} & \sethlcolor{cellcolor11}\hl{57.4} & \sethlcolor{cellcolor12}\hl{42.4} & \sethlcolor{cellcolor73}\hl{34.0} & \sethlcolor{cellcolor74}\hl{93.4} & \sethlcolor{cellcolor75}\hl{70.9} & \sethlcolor{cellcolor76}\hl{58.1} & \sethlcolor{cellcolor77}\hl{53.3} & \sethlcolor{cellcolor78}\hl{31.6} \\
& vit\_gpt2 & \sethlcolor{cellcolor13}\hl{1.2} & \sethlcolor{cellcolor14}\hl{5.6} & \sethlcolor{cellcolor15}\hl{3.2} & \sethlcolor{cellcolor16}\hl{9.7} & \sethlcolor{cellcolor79}\hl{6.1} & \sethlcolor{cellcolor80}\hl{15.1} & \sethlcolor{cellcolor81}\hl{31.6} & \sethlcolor{cellcolor82}\hl{59.3} & \sethlcolor{cellcolor83}\hl{18.2} & \sethlcolor{cellcolor84}\hl{129.9} \\
&BLIP & \sethlcolor{cellcolor17}\hl{-5.2} & \sethlcolor{cellcolor18}\hl{0.6} & \sethlcolor{cellcolor19}\hl{-3.2} & \sethlcolor{cellcolor20}\hl{1.8} & \sethlcolor{cellcolor85}\hl{1.9} & \sethlcolor{cellcolor86}\hl{1.1} & \sethlcolor{cellcolor87}\hl{7.2} & \sethlcolor{cellcolor88}\hl{5.7} & \sethlcolor{cellcolor89}\hl{6.7} & \sethlcolor{cellcolor90}\hl{16.8} \\
\midrule
&ClipCap & \sethlcolor{cellcolor21}\hl{21.3} & \sethlcolor{cellcolor22}\hl{10.3} & \sethlcolor{cellcolor23}\hl{30.9} & \sethlcolor{cellcolor24}\hl{16.6} & \sethlcolor{cellcolor91}\hl{13.7} & \sethlcolor{cellcolor92}\hl{20.4} & \sethlcolor{cellcolor93}\hl{37.6} & \sethlcolor{cellcolor94}\hl{115.7} & \sethlcolor{cellcolor95}\hl{10.3} & \sethlcolor{cellcolor96}\hl{30.8} \\
&mPLUG & \sethlcolor{cellcolor25}\hl{30.9} & \sethlcolor{cellcolor26}\hl{28.6} & \sethlcolor{cellcolor27}\hl{55.5} & \sethlcolor{cellcolor28}\hl{33.9} & \sethlcolor{cellcolor97}\hl{31.1} & \sethlcolor{cellcolor98}\hl{24.5} & \sethlcolor{cellcolor99}\hl{79.0} & \sethlcolor{cellcolor100}\hl{294.9} & \sethlcolor{cellcolor101}\hl{213.2} & \sethlcolor{cellcolor102}\hl{50.0} \\
Flickr30k & GIT & \sethlcolor{cellcolor29}\hl{45.2} & \sethlcolor{cellcolor30}\hl{32.5} & \sethlcolor{cellcolor31}\hl{19.2} & \sethlcolor{cellcolor32}\hl{27.6} & \sethlcolor{cellcolor103}\hl{24.0} & \sethlcolor{cellcolor104}\hl{155.4} & \sethlcolor{cellcolor105}\hl{19.0} & \sethlcolor{cellcolor106}\hl{45.7} & \sethlcolor{cellcolor107}\hl{87.9} & \sethlcolor{cellcolor108}\hl{-12.3} \\
& vit\_gpt2 & \sethlcolor{cellcolor33}\hl{20.8} & \sethlcolor{cellcolor34}\hl{17.1} & \sethlcolor{cellcolor35}\hl{34.9} & \sethlcolor{cellcolor36}\hl{25.9} & \sethlcolor{cellcolor109}\hl{19.0} & \sethlcolor{cellcolor110}\hl{117.1} & \sethlcolor{cellcolor111}\hl{76.1} & \sethlcolor{cellcolor112}\hl{121.4} & \sethlcolor{cellcolor113}\hl{35.0} & \sethlcolor{cellcolor114}\hl{160.3}  \\
&BLIP & \sethlcolor{cellcolor37}\hl{-6.5} & \sethlcolor{cellcolor38}\hl{1.3} & \sethlcolor{cellcolor39}\hl{-2.2} & \sethlcolor{cellcolor40}\hl{1.6} & \sethlcolor{cellcolor115}\hl{1.5} & \sethlcolor{cellcolor116}\hl{2.2} & \sethlcolor{cellcolor117}\hl{5.5} & \sethlcolor{cellcolor118}\hl{8.3} & \sethlcolor{cellcolor119}\hl{-3.6} & \sethlcolor{cellcolor120}\hl{14.5} \\
\midrule
&ClipCap & \sethlcolor{cellcolor41}\hl{14.6} & \sethlcolor{cellcolor42}\hl{8.4} & \sethlcolor{cellcolor43}\hl{21.3} & \sethlcolor{cellcolor44}\hl{14.0} & \sethlcolor{cellcolor121}\hl{12.6} & \sethlcolor{cellcolor122}\hl{15.1} & \sethlcolor{cellcolor123}\hl{27.0} & \sethlcolor{cellcolor124}\hl{122.0} & \sethlcolor{cellcolor125}\hl{0.0} & \sethlcolor{cellcolor126}\hl{17.8} \\
&mPLUG & \sethlcolor{cellcolor45}\hl{148.6} & \sethlcolor{cellcolor46}\hl{49.8} & \sethlcolor{cellcolor47}\hl{60.3} & \sethlcolor{cellcolor48}\hl{40.7} & \sethlcolor{cellcolor127}\hl{36.8} & \sethlcolor{cellcolor128}\hl{53.1} & \sethlcolor{cellcolor129}\hl{79.6} & \sethlcolor{cellcolor130}\hl{$\infty$} & \sethlcolor{cellcolor131}\hl{143.9} & \sethlcolor{cellcolor132}\hl{62.0} \\
XM3600 & GIT & \sethlcolor{cellcolor49}\hl{46.5} & \sethlcolor{cellcolor50}\hl{23.9} & \sethlcolor{cellcolor51}\hl{23.6} & \sethlcolor{cellcolor52}\hl{11.9} & \sethlcolor{cellcolor133}\hl{11.4} & \sethlcolor{cellcolor134}\hl{4.7} & \sethlcolor{cellcolor135}\hl{13.3} & \sethlcolor{cellcolor136}\hl{70.8} & \sethlcolor{cellcolor137}\hl{33.7} & \sethlcolor{cellcolor138}\hl{1.3} \\
&vit\_gpt2 & \sethlcolor{cellcolor53}\hl{32.3} & \sethlcolor{cellcolor54}\hl{18.3} & \sethlcolor{cellcolor55}\hl{34.1} & \sethlcolor{cellcolor56}\hl{21.8} & \sethlcolor{cellcolor139}\hl{17.6} & \sethlcolor{cellcolor140}\hl{77.3} & \sethlcolor{cellcolor141}\hl{67.8} & \sethlcolor{cellcolor142}\hl{89.3} & \sethlcolor{cellcolor143}\hl{72.7} & \sethlcolor{cellcolor144}\hl{89.1} \\
&BLIP & \sethlcolor{cellcolor57}\hl{-3.5} & \sethlcolor{cellcolor58}\hl{1.9} & \sethlcolor{cellcolor59}\hl{1.8} & \sethlcolor{cellcolor60}\hl{0.7} & \sethlcolor{cellcolor145}\hl{0.9} & \sethlcolor{cellcolor146}\hl{-1.6} & \sethlcolor{cellcolor147}\hl{3.7} & \sethlcolor{cellcolor148}\hl{-7.6} & \sethlcolor{cellcolor149}\hl{4.6} & \sethlcolor{cellcolor150}\hl{6.0} \\
\bottomrule
\end{tabular}%
}
\caption{Performance change after reformulation compared to raw model output on common metrics, datasets and models (in \% of the recorded performance before reformulation). We observe major improvements in weaker models (ClipCap, mPLUG, GIT, vit\_gpt2). Darker green (red) indicates higher improvement (deterioration). M: METEOR, C: CIDEr, S: SPICE. $\infty$ marks a configuration where the metric value before reformulation was 0.}
\label{tab:re_during_inference}
\end{table*}

%% file: sections/04.3_german.tex
The last section demonstrated strong gains of our approach for weak off-the-shelf models. Acknowledging that image captioning models sharply drop in performance in languages other than English, we next investigate the use of English reformulation in a cross-lingual setup. We combine a German image captioning model with our reformulation model by generating German captions; translating the captions to English; reformulating them with our model; and translating back to German.

\paragraph{Data.}
We use Multi30k~\cite{elliott2016multi30k}, a large, non translated, German image caption dataset, 
which contains 30K/1K images for train/test, each with 5 captions. All images are taken from the Flickr30k dataset and all captions are generated by German native speakers.

\paragraph{Model.}
Due to a lack of a strong and publicly available pretrained image captioning model for German, we train our own model. We use the ClipCap model as it separates the text decoder from the image encoder, allowing us to straighforwardly incorporate a German decoder. We use the original ClipCap implementation\footnote{\href{https://github.com/rmokady/CLIP_prefix_caption}{github.com/rmokady/CLIP\_prefix\_caption}} and change the text decoder to a German version of GPT2.\footnote{\href{https://huggingface.co/dbmdz/german-gpt2}{huggingface.co/dbmdz/german-gpt2}} We refer to this model as \textbf{base}. We reformulate the captions generated by \textbf{base}, and refer to these as \textbf{base+re}. Following recent captioning works~\citep{thapliyal2022crossmodal, ramos2023lmcap}, we use Google Translation API for all translations.


\paragraph{Baselines.}
First, to directly measure the performance gain of the reformulation pipeline, we use \textbf{base} as a baseline.
Second, the mPLUG checkpoint on which the reformulation model is based (see Section~\ref{sec:reformulation_feedback}) is in itself quite a capable captioning model. Consequently, given an input image and caption the reformulation model might ignore the input caption and generate its own caption.
To make sure this is not the case, we also generate English captions using the reformulation model by providing an image and an empty caption as input, and translate these captions to German (\textbf{tran}).
Finally, we present results reported by recent German image captioning studies: Dual Attention~\citep[\textbf{DA},][]{jaffe2017generating}, Cycle Consistency~\citep[\textbf{CC},][]{wu2019improving} and Multi-Objective Optimization~\citep[\textbf{MOO},][]{wu2022pairs}. We report the same metrics as in Section~\ref{sec:english_auto_eval} except SPICE which, to the best of our knowledge, is not available for German.

\input{tables_and_plots/experiments/german_multi30k_results}

\paragraph{Results.}
Results in Table~\ref{tab:german_multi30k_results} show that \textbf{base+re} outperforms all other methods in BLEU-4 and CIDEr, while \textbf{tran} achieve the best result in METEOR, though by a small margin. The improvement over \textbf{base} emphasizes the power of the reformulation pipeline, while the improvement over \textbf{tran} suggests that providing the reformulation model with a reasonable caption is an important factor in the success of the reformulation process. We also note the improvement over previous state-of-the-art studies. We partially attribute this to the use of the strong German GPT2 model (since \textbf{base} outperforms previous models on two metrics), but reformulation contributes notable value, as evidenced by the superiority of \textbf{base+re} over \textbf{base}.

\subsubsection{Human Evaluation}

To better understand the improvement reported by the automatic metrics, we follow the same protocol as in Section~\ref{sec:english_human_eval}. The annotation was conducted by two on-site German native speakers with an inter-annotator agreement score (measured by Cohen's Kappa) of at least 0.54 across all axes.

\begin{figure} [tb]
    \centering
    \includegraphics[width=\columnwidth, height=3.5cm]{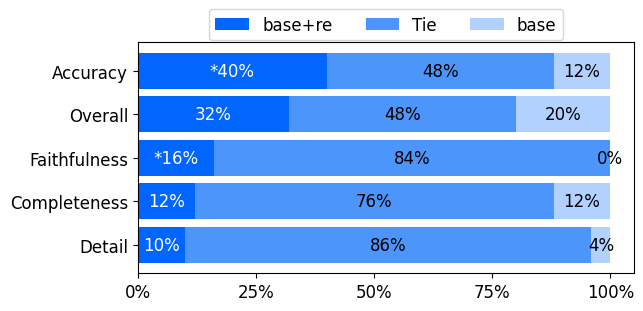}
    \caption{
    Results for human evaluation on different axes of German generated captions. We show proportions of preferences for generated captions without (base) and with (base+re) reformulations, and ties.
    * indicates significance as in Figure~\ref{fig:english_human_eval}.
    }
    \label{fig:german_human_eval}
\end{figure}

Results are presented in Figure~\ref{fig:german_human_eval}. We notice that while in English improvement was most significant in terms of Completeness (Figure~\ref{fig:english_human_eval}), in German the most significant axes are Faithfulness and Accuracy. We hypothesize that the captions produced by the German \text{base} model contain many errors and the focus of the reformulation process is therefore on fixing the errors, while errors in the English generated captions are rare and thus the focus is on adding new information. We corroborate this hypothesis by computing the mean caption length before and after reformulation
for English (44.6 $\rightarrow$ 49.3) and German (57.3 $\rightarrow$ 54.1).
See Appendix~\ref{sec:app_more_examples} for examples.

%% file: tables_and_plots/experiments/german_multi30k_results.tex
\begin{table}[!t]
\small
\centering
\resizebox{6.42cm}{!}{%
\begin{tabular}{cccc}
\toprule
 & B@4 & METEOR & CIDEr \\
\midrule
base & 12.8 $\pm$ 0.3 & 18.6 $\pm$ 0.2 & 39.2 $\pm$ 1.6 \\
tran & 14.3 & \textbf{20.3} & 46.0 \\
\midrule
DA & 16.0 & 17.8 & 30.8 \\
CC & 15.9 & 17.8 & 31.0 \\
MOO & 16.5 & 17.9 & 33.8 \\
\midrule
base+re & \textbf{16.8 $\pm$ 0.1} & 20.1 $\pm$ 0.1 & \textbf{51.4 $\pm$ 0.6} \\
\bottomrule
\end{tabular}%
}
\caption{German Results on Multi30k test set. Results for the models that we train (base, base+re) are averaged over 3 random initializations and we report the standard deviation. For each metric, the best result is bolded.}
\label{tab:german_multi30k_results}
\end{table}

%% file: sections/05_stylized_captioning.tex
We study the generalizability of reformulation feedback modeling by focusing on a second reformulation attribute: the style of the caption, i.e., the reformulation should adapt the style while making minimal changes.

\subsection{Dataset}
We use the FlickrStyle~\citep{gan2017stylenet} dataset. FlickrStyle contains humorous and romantic captions for 7000 images from Flickr30K. Importantly, the annotators were instructed to generate the captions based on existing captions from Flickr30K. We follow \citet{wang2023tridentcap} and randomly split the data to 6000 train images and 1000 test images.

\subsection{Method} \label{sec:style_transfer_re_model}
We train a reformulation model for a given style as follows. First, for each caption in FlickrStyle we identify the original caption in Flickr30K on which that caption is based by measuring the string overlap of the stylized caption with each of the original captions of the same image, and selecting the caption with the largest overlap. Next, we fine-tune a reformulation model as described in Section~\ref{sec:reformulation_feedback}, with the original caption as the input and the stylized caption as the ground-truth output.

\subsection{Models}
We use BLIP as the captioning model (\textbf{BLIP)} and for each style, we reformulate the BLIP captions using a reformulation model trained to transfer captions to the style in question (\textbf{BLIP+re}). Note that vanilla BLIP does not generate stylized captions (i.e., is expected to perform poorly on this task). As baselines, we present results from previous studies: CapDec~\citep{nukrai2022text}, SAN~\citep{li2021similar}, and TridentCap~\citep{wang2023tridentcap}.

\input{tables_and_plots/experiments/style_results}

\subsection{Automatic evaluation}
\begin{figure} [tb]
    \centering
    \includegraphics[width=\columnwidth, height=3.7cm]{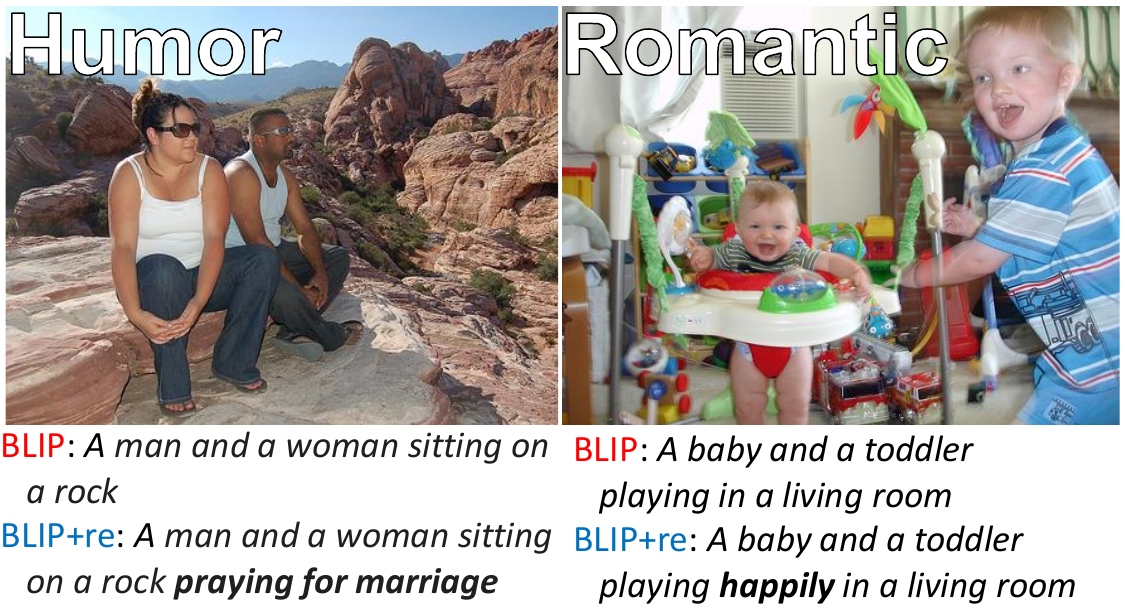}
    \caption{Examples of original captions and reformulated captions for humor and romantic reformulation.}
    \label{fig:style_examples}
\end{figure}

Results are presented in Table~\ref{tab:style_results}. We follow the convention from previous stylized image captioning studies and report Bleu-1, Bleu-3, METEOR and CIDEr. Our method achieves state-of-the-art results for both styles, and we attribute this improvement to the strong captions generated by the BLIP model (in the humor style \textbf{BLIP} even outperforms \textbf{BLIP+re} in the CIDEr metric).
This unveils an issue in automatic evaluation: vanilla BLIP outperformed the baselines though it clearly does not generate stylized captions (see Figure~\ref{fig:style_examples} for examples). The same may be true for \textbf{BLIP+re}. Thus, we conduct human evaluation to ensure that captions generated by \textbf{BLIP+re} are indeed stylized.

\subsection{Human Evaluation}

We again use our human evaluation scheme (Section~\ref{sec:english_human_eval}) to compare to previous baselines. We compare to CapDec\footnote{\href{https://github.com/DavidHuji/CapDec}{github.com/DavidHuji/CapDec}}, since we found no available codebases for TridentCap and SAN. We ask the first 4 questions from Section~\ref{sec:english_human_eval} (Faithfulness, Completness, Accuracy, Detail) and add a style-related question: \emph{Which caption is more \{humorous,romantic\}?}

\begin{figure} [tb]
    \centering
    \includegraphics[width=\columnwidth, height=5.3815cm]{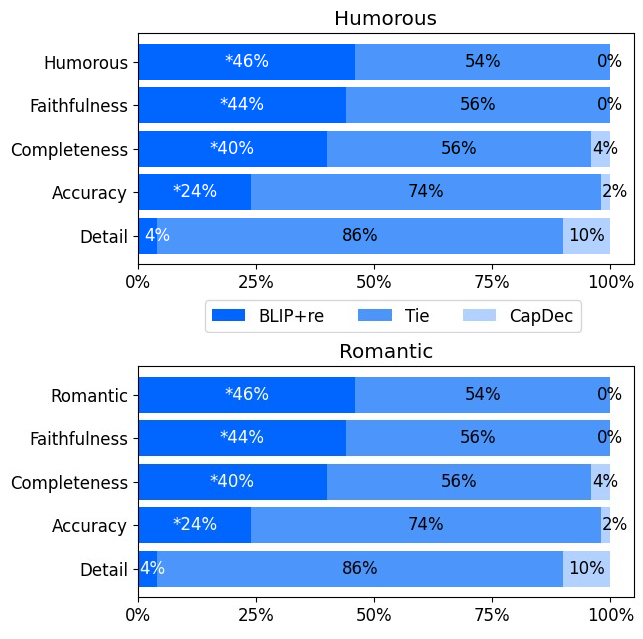}
    \caption{
    Results for human evaluation on different axes of stylized captions. We show proportions of preferences for the baseline (CapDec) and BLIP reformulated (BLIP+re) captions, and ties.
    * indicates significance as in Figure~\ref{fig:english_human_eval}.}
    \label{fig:style_human_eval}
\end{figure}

Results are presented in Figure~\ref{fig:style_human_eval}. Our method improves over the baseline not only in the quality of captions, but also in generating stylized captions, significantly in both styles. Annotators agreement (Fleiss' Kappa) values were $\kappa = 0.59, 0.51, 0.48, 0.34$ for Faithfulness, Style, Completeness, Accuracy (the axes where reformulated captions were better), and $\kappa = 0.44$ for Detail.


%% file: tables_and_plots/experiments/style_results.tex
\begin{table}[!t]
\small
\centering
\resizebox{7.3cm}{!}{%
\begin{tabular}{cccccc}
\toprule
Style & Method & B@1 & B@3 & M & C \\
\midrule
\multirow{5}{*}{Humorous}
& CapDec & 29.4 & 8.8 & 13.2 & 55.1 \\
& SAN & 29.5 & 9.9 & 12.5 & 47.2 \\
& TridentCap & 30.6 & 11.2 & 12.8 & 56.6 \\
\cmidrule{2-6}
& BLIP & 29.6 & 11.0 & 14.4 & \textbf{73.9} \\
& BLIP+re & \textbf{33.7} & \textbf{11.7} & \textbf{14.8} & 72.0 \\
\midrule
\multirow{5}{*}{Romantic}
& CapDec & 27.9 & 8.9 & 12.6 & 52.2 \\
& SAN & 30.9 & 10.9 & 13.0 & 53.3 \\
& TridentCap & 31.9 & 11.4 & 13.4 & 60.4 \\
\cmidrule{2-6}
& BLIP & 28.5 & 11.2 & 14.3 & 72.0 \\
& BLIP+re & \textbf{35.1} & \textbf{13.0} & \textbf{15.4} & \textbf{74.6} \\
\bottomrule
\end{tabular}%
}
\caption{Results for stylized image captioning on FlickrStyle. B@n: BLEU-n, M: METEOR, C: CIDEr. For each style and metric, the best result is in bold.}
\label{tab:style_results}
\end{table}

%% file: sections/06_discussion.tex

Despite the recent success of incorporating (models of) human feedback as a training signal, using feedback during inference has received little attention. We presented a novel approach -- reformulation feedback at inference time -- and applied it to the task of image captioning.

We refrain from comparing our approach to a baseline of fine-tuning the captioning model directly on the corrected captions for two reasons. First, even if such baseline would induce better results, our method's advantage is efficiency, as reformulation models are trained once and can be combined with any base model architecture, while fine-tuning would be performed on any new model. Second, this baseline is not applicable in our cross-lingual use-case (Section~\ref{sec:german}), as the corrected captions are in English.

We've studied two use-cases for our method: improving captioning models in challenge domains (Section~\ref{sec:german}) and generating high quality stylized captions (Section~\ref{sec:style_transfer}).
Both can be extended in future work: captioning models in other challenge domains (e.g., medical image captioning) can gain improved factuality, while robust models can be utilized to generate captions in other styles (e.g., sentimental captions).
Taken together, our work contributes to the active areas of learning from human feedback, and efficient adaptation of powerful LLMs to diverse tasks.


%% file: sections/limitations.tex
\paragraph{Data collection.}
While our method requires less computational resources compared to previous studies (since only the feedback model is trained rather than the generative model), it requires more human resources for annotation. Simpler types of feedback (e.g., the common comparative feedback) require less effort and time per sample than reformulation, while some studies~\citep[e.g.][]{ramos2023aligning} refrain from explicitly collecting any feedback data, by using publicly available human annotations that were originally collected for a different purpose (e.g., to train evaluation metrics).

\paragraph{Cross-lingual reformulation.}
The pipeline suggested in Section~\ref{sec:german} for cross-lingual reformulation (generation of captions in the target language, translation into English, reformulation, translation back into the target language) depends on the existence of a decent base captioning model in the target language and good translation models from/to English. If the base captioning model in the target language generates poor captions, the reformulated captions will be no better than captions generated in English and translated to the target language (i.e. the \textbf{tran} baseline discussed in Section~\ref{sec:german}). If there are no strong translation models from/to English, the quality of captions would decrease in every translation step in the pipeline, resulting in poor captions. Future work may address training non-English reformulation model to bridge the second gap.

\paragraph{Variation in annotation conditions.}
Previous studies~\citep{khashabi-etal-2022-genie} show that human annotations may vary drastically when basic conditions change, e.g., on different days or even at a different time during the day. Since reformulation models are trained on such annotations, this may have a significant impact on the model. We did not take this into account in our data collection and usage.

%% file: sections/ethics_statement.tex
In our data collection in Section~\ref{sec:data_collection} we collect no identifying data on the annotators. For existing datasets, we use publicly available resources in accordance with their license agreements. The datasets are fully anonymized and do not contain personal information about the caption annotators or any information that could reveal the identity of the photographed subjects.

As with other methods for modifying model outputs, our approach can be used to transfer toxic text to non-toxic text, or vice versa. Additionally, the reformulation data that was collected and presented in Section~\ref{sec:error_correction} may contain social biases. Along with the publication of our model and data, we will include a model card~\citep{mitchell2019model} which reports standard information regarding the collected data, training methods and intended use.

This work was approved by the The School of Computer Science \& Engineering Committee for the Use of Human Subjects in Research in the Hebrew University of Jerusalem.

%% file: sections/app_a_training_details.tex
\subsection{Reformulation Models}

We now specify the details of the reformulation models trained in Sections~\ref{sec:error_correction} and \ref{sec:style_transfer_re_model}.

We use the VQA training pipeline from the official mPLUG code base.\footnote{\href{https://github.com/alibaba/AliceMind}{github.com/alibaba/AliceMind}}
We use the default hyperparameters, and fine-tune the mplug.en.base checkpoint for 8 epochs with the AdamW optimizer and learning rate of 3e-5. Models were trained on an Nvidia RTX a5000 GPU and each training session took less than an hour. Models contain 350M parameters.

\subsection{German Captioning Model}

We train the model discussed in Section~\ref{sec:german} for 10 epochs with the AdamW optimizer and learning rate of 2e-5. The model was trained on an Nvidia RTX a5000 GPU and training took 4 hours to complete. The model contains 156M parameters.

%% file: sections/app_b_data_collection.tex
In this section we thoroughly discuss the data collection process briefly discussed in Section~\ref{sec:data_collection}.

We use Amazon Mechanical Turk to recruit annotators. As a first filter we require native-speaker level proficiency in English. Next, we publish a qualification task and filter the annotators. Finally, after each batch of annotation, we sample 20 annotated samples to ensure the quality of annotations, and inform annotators if a wrong annotation has been made.

Annotators were paid 0.1\$US per annotation. Early experiments indicated that a single reformulation annotation takes 5 to 30 seconds. The expected hourly wage exceeds the US minimum wage which ranges between 8\$US and 15\$US.

We provide the following annotation guidelines:
\begin{itemize}
    \item In this task, you will be presented with images together with a textual image description.
    \item Your task is to reformulate the description so that (a) it is as similar as possible to the original (b) all errors from the original descriptions are fixed (if any errors exist).
    \item If the original description is too bad to fix, please write a completely new description.
\end{itemize}

Subsequently, annotators were shown several examples of reformulations.


%% file: sections/app_c_used_packages.tex
We used the following packages in our implementation:

\begin{itemize}
    \item COCO-caption evaluation\footnote{\href{https://github.com/tylin/coco-caption}{github.com/tylin/coco-caption}}: used for all evaluation metrics.
    \item statsmodel: used for sign-test\footnote{\href{https://www.statsmodels.org/stable/generated/statsmodels.stats.descriptivestats.sign_test.html}{www.statsmodels.org/stable/generated/statsmodels.stats.descriptivestats.\\sign\_test.html}} and Fleiss' Kappa\footnote{\href{https://www.statsmodels.org/stable/generated/statsmodels.stats.inter_rater.fleiss_kappa.html}{https://www.statsmodels.org/stable/generated/statsmodels.stats.inter\_rater.\\fleiss\_kappa.html}} in the human evaluation sections.
    \item sklearn: used for Cohen's Kappa\footnote{\href{https://scikit-learn.org/stable/modules/generated/sklearn.metrics.cohen_kappa_score.html}{scikit-learn.org/stable/modules/generated/sklearn.metrics.\\cohen\_kappa\_score.html}} in Section~\ref{sec:german}.
\end{itemize}

%% file: sections/app_d_more_examples.tex
\begin{figure*} [tb]
    \centering
    \includegraphics[width=\textwidth]{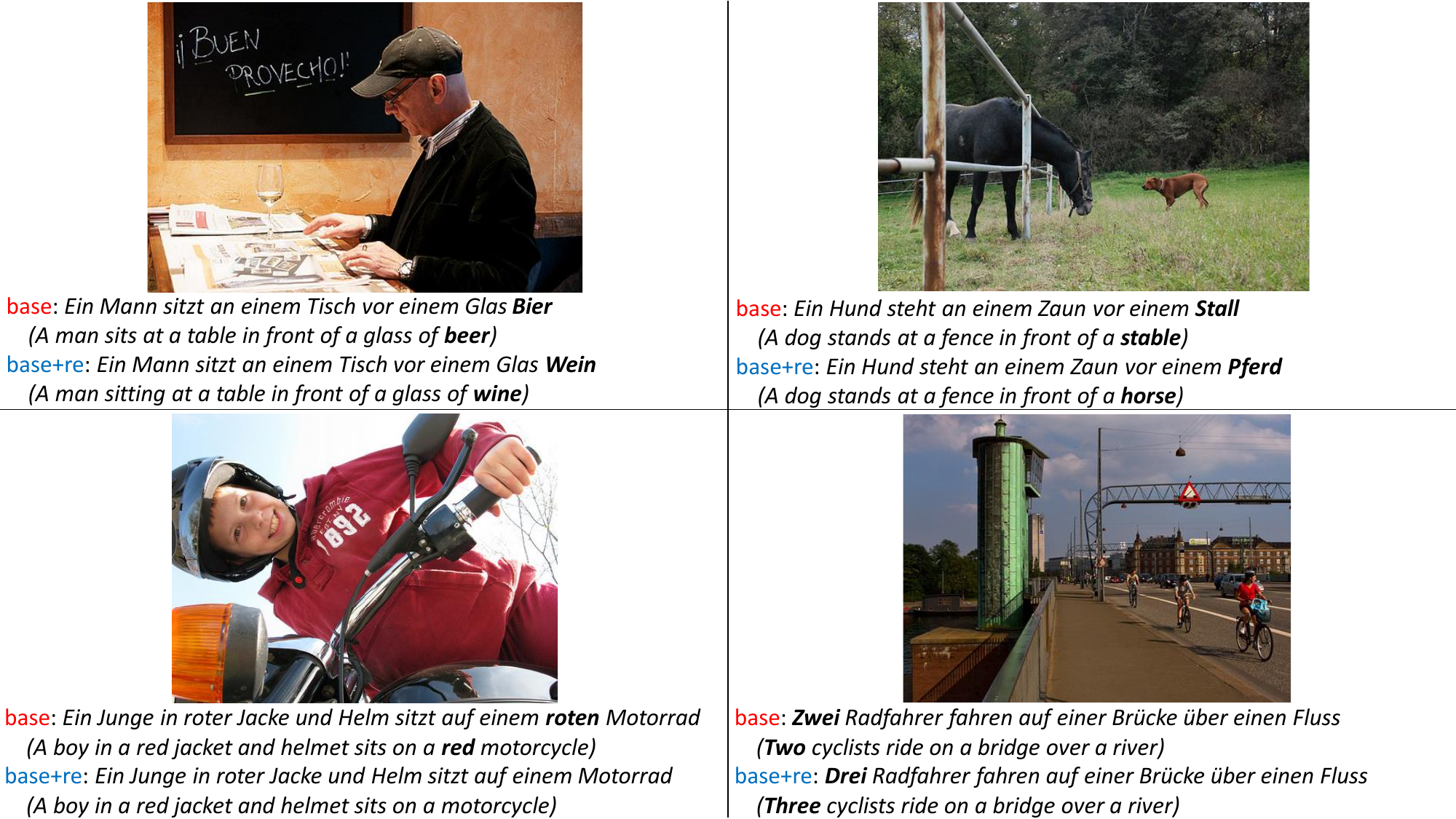}
    \caption{Examples in which the reformulated captions fix errors in captions generated by the base model, for German image captioning. base: the caption generated by the base model. base+re: the reformulated caption.}
    \label{fig:german_improvement}
\end{figure*}

Figure~\ref{fig:german_improvement} presents samples where the German captioning \textbf{base} model discussed in Section~\ref{sec:german} generates caption with errors, which are fixed by the reformulation process.

%% file: sections/app_e_blip_analysis.tex
We use the evaluation framework described in Section~\ref{sec:english_human_eval} on the BLIP model. We randomly sample 50 images from each of the MSCOCO and Flickr30k test sets for the evaluation.

\begin{figure} [tb]
    \centering
    \includegraphics[width=\columnwidth]{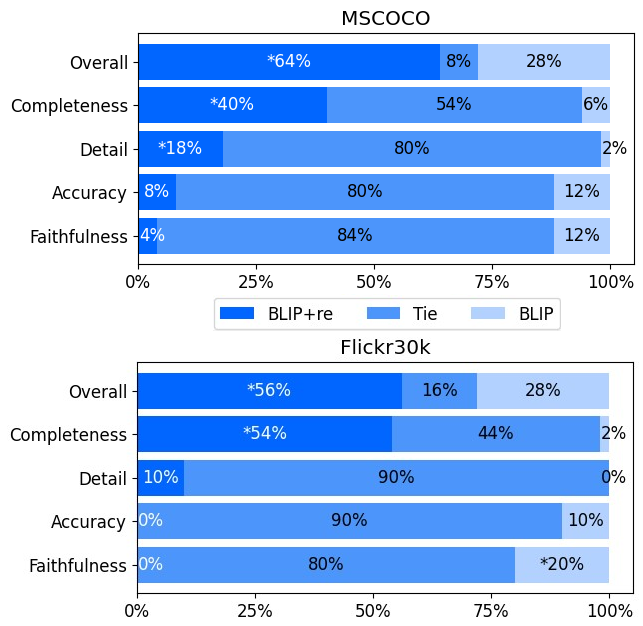}
    \caption{Results for human evaluation on BLIP in different axes. We show proportions of preferences for generated captions without (base) and with (base+re) reformulations, and ties.
    * indicates a significant difference between base and base+re (Sign test; $p<0.05$).
    }
    \label{fig:blip_human_eval}
\end{figure}

Figure~\ref{fig:blip_human_eval} presents the results. Across datasets, reformulated captions are more complete and detailed but less faithful and accurate. This result is in line with the analysis presented in Table~\ref{tab:re_dataset_analysis}, where the most common feedback type was `addition' of information to the original caption. The reduction in accuracy and faithfulness shows that in some cases the added information was incorrect. However, annotators scored the reformulated captions as overall better in both datasets.

We find that reformulated captions are significantly (Sign test, $p < 0.05$) more detailed in MSCOCO, less faithful in Flickr30k, more complete in both datasets and overall better in both datasets ($p<0.05$). We also compute inter-annotator agreement using Fleiss' Kappa: $\kappa = 0.55, 0.47, 0.44$ for completeness, overall, detail (axes on which reformulated captions were better), and $\kappa = 0.37, 0.34$ for faithfulness, accuracy.